\documentclass[10pt,twocolumn,letterpaper]{article}

\usepackage{wacv}
\usepackage{times}
\usepackage{epsfig}
\usepackage{graphicx}
\usepackage{amsmath}
\usepackage{amssymb}
\usepackage{booktabs}
\usepackage{dirtytalk}
\usepackage{amssymb}
\usepackage{mathtools}
\usepackage{booktabs}

\usepackage[normalem]{ulem}

\def\wacvPaperID{36} 

\wacvfinalcopy 

\ifwacvfinal
\def\assignedStartPage{9876} 
\fi

\usepackage{epsfig}%
\usepackage{graphicx}
\usepackage{comment}
\usepackage{amsmath,amssymb} 
\usepackage{color}
\usepackage{dirtytalk}%
\usepackage{amssymb}%
\usepackage{mathtools}%
\usepackage{sidecap}
\usepackage[normalem]{ulem}%
\usepackage{multirow}
\usepackage{subfigure}

\usepackage[useames,dvipsnames,svgnames,table]{xcolor}
\usepackage{xspace}
\makeatletter
\DeclareRobustCommand\onedot{\futurelet\@let@token\@onedot}
\def\@onedot{\ifx\@let@token.\else.\null\fi\xspace}
\def\eg{\emph{e.g}\onedot} 

\def\ie{\emph{i.e}\onedot}

\def\wrt{w.r.t\onedot} 

\def\etal{\emph{et al}\onedot}
\makeatother

\def\Vec#1{{\boldsymbol{#1}}}
\def\Mat#1{{\boldsymbol{#1}}}

\DeclareMathOperator*{\argmin}{arg\,min}

\usepackage{float}

\usepackage{algorithm}
\usepackage{algpseudocode}
\usepackage{amsmath}

\ifwacvfinal
\usepackage[breaklinks=true,bookmarks=false]{hyperref}
\else
\usepackage[pagebackref=true,breaklinks=true,colorlinks,bookmarks=false]{hyperref}
\fi

\ifwacvfinal
\else
\fi

\begin{document}

\title{Set Augmented Triplet Loss for Video Person Re-Identification}


\author{Pengfei Fang$^{1,2}$, Pan Ji$^3$\thanks{Work done while at NEC Laboratories America}~, Lars Petersson$^{2}$, Mehrtash Harandi$^{4}$ \\
$^{1}$Australian National University, $^{2}$DATA61-CSIRO, $^3$OPPO US Research Center, $^{4}$Monash University\\

{\tt\small Pengfei.Fang@anu.edu.au, peterji1990@gmail.com}\\
{\tt\small Lars.Petersson@data61.csiro.au, mehrtash.harandi@monash.edu}
}

\maketitle

\begin{abstract}
Modern video person re-identification (re-ID) machines are often trained using a metric learning approach, supervised by a triplet loss. The triplet loss used in video re-ID is usually based on so-called clip features, each aggregated from a few frame features. In this paper, we propose to model the video clip as a set and instead study the distance between sets in the corresponding triplet loss. In contrast to the distance between clip representations, the distance between clip sets considers the pair-wise similarity of each element (\ie, frame representation) between two sets. This allows the network to directly optimize the feature representation at a frame level. Apart from the commonly-used set distance metrics (\eg, ordinary distance and Hausdorff distance), we further propose a hybrid distance metric, tailored for the set-aware triplet loss. Also, we propose a hard positive set construction strategy using the learned class prototypes in a batch. Our proposed method achieves state-of-the-art results across several standard benchmarks, demonstrating the advantages of the proposed method.
\end{abstract}



\section{Introduction}\label{sec:intro}


Person re-identification (re-ID) has drawn an increasing amount of attention in academia and industry due to its great potential in research and real-world applications~\cite{ZhengLiang2016arXivReIDPastPresentFuture}. A person re-ID machine is trained to regress a non-linear function which maps the pedestrian images to a semantically meaningful embedding space. In such an embedding space, feature vectors extracted from images belonging to the same identity (ID) are clustered, thereby retrieving correct matches for unseen query images of persons in the gallery. In the past decades, image re-ID has achieved significant improvements via learning discriminative representations from a single image~\cite{Suh_2018_ECCV, Sun_2018_ECCV, Pengfei_2019_ICCV,TBCL}. Recently, video person re-ID has attracted a growing interest as videos provide richer cues in terms of encoding video representations for person retrieval~\cite{Li_2019_ICCV,Zhao_2019_CVPR,Fu2018STASA,COSAM_2019_ICCV,Fang_2020_ACCV}. In this paper, we aim to create compact yet discriminative features from videos for accurate video re-ID.



\begin{figure*}[t]
\centering
	\subfigure[]{\includegraphics[width=0.32\linewidth,height=3.8cm]{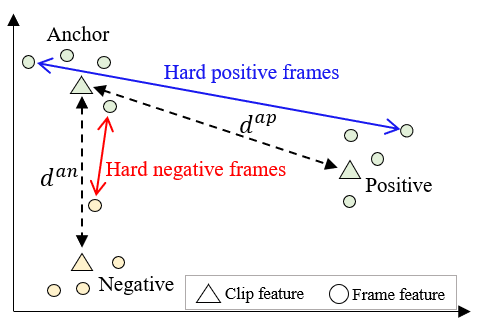}\label{fig:example}}%
	\hfil
	\subfigure[]{\includegraphics[width=0.32\linewidth,height=3.8cm]{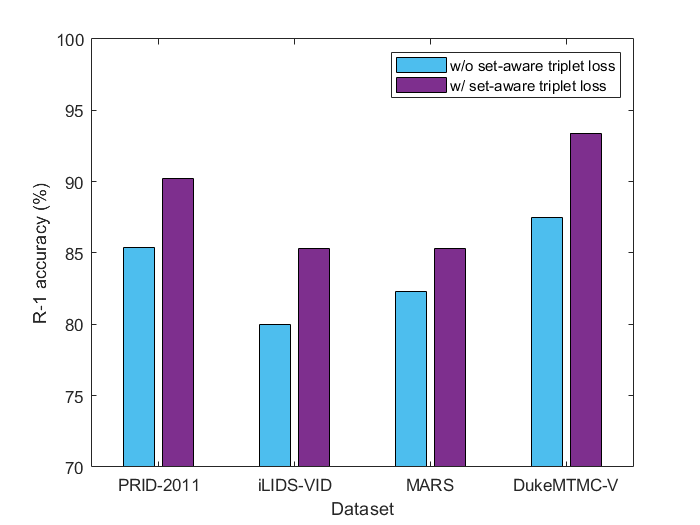}\label{fig:bar}}%
	\hfil
	\caption{(a): Geometry interpretation of the distance metrics for clip representation and frame representation. The color represents the class of samples. $d^{ap}$ and $d^{an}$ denote the distance from positive pair and negative pair in a clip level. However, those two distances cannot reveal the original distribution of frame features, thereby ignoring the distance between hard frames (\ie, \textcolor{red}{$\leftrightarrow$} for hard negative pair and \textcolor{blue}{$\leftrightarrow$} for hard positive pair). (b): The comparison of R-1 accuracy from the networks trained without set-aware triplet loss and with set-aware triplet loss, across four datasets. The backbone network is ResNet-50, pre-trained on ImageNet. In the set-aware triplet loss, we use the proposed hybrid set distance metric to calculate the distance of anchor-positive pair and anchor-negative pair.}
	\label{fig:illustration}
\end{figure*}

The pipeline of training a typical video re-ID machine consists of first extracting the frame-level features with the help of a deep network backbone and then aggregating them to a clip-level feature. 
In video re-ID, the ranking task (\ie, triplet loss) is a popular choice to supervise the network to learn an embedding space, \wrt the clip-level features. 
This, however, could lead to sub-optimal learning of the video embedding space, as the aggregation operation to frame features will result in loss of information of the original frame features. Specifically, in the video-based applications, the triplet loss considers the distance between the clip representations (\ie, $d^{an}$ and $d^{ap}$ in Fig.~\ref{fig:example}), which only indirectly penalizes the hard frames between the clips (\ie, hard positive frames and hard negative frames in Fig.~\ref{fig:example}).
This observation motivates us to directly leverage the frame features, to decrease the hard positive distance (\ie, \textcolor{blue}{$\leftrightarrow$} in Fig.~\ref{fig:example}) and increase the hard negative distance (\ie, \textcolor{red}{$\leftrightarrow$} in Fig.~\ref{fig:example}) for frame features.



In video re-ID, we often aggregate the frame features (\ie,  $\{\Vec{f}_1, \ldots, \Vec{f}_t\},\Vec{f}_i \in \mathbb{R}^{c}, i = 1, \ldots, t$) to a clip-level representation (\ie, ${\Vec{\hat{f}}} \in \mathbb{R}^c$) using an aggregation function (\ie, $\mathrm{Agg(\cdot)}$). This processing can be summarized as: 
\begin{equation}\label{eq:fusion}
{\Vec{\hat{f}}} = \mathrm{Agg}(\{\Vec{f}_1, \ldots, \Vec{f}_t\}) = \phi\big(\sum_{i = 1}^t(\omega_i\Mat{f}_i)\big),  
\end{equation}
where $\phi(\cdot)$ and $\{\omega_1, \ldots, \omega_t\} \in \mathbb{R}^t$ denote non-linear mapping and aggregation weights, respectively. Due to the summation operator in Eqn.~\eqref{eq:fusion}, the clip feature (\ie, ${\Vec{\hat{f}}}$) is invariant to the order of frame features, indicating that the aggregation function is temporally invariant. In other words, the aggregation function acts on sets, in the sense that the response of the aggregation function is \say{insensitive} to the ordering of elements in the input~\cite{NIPS2017_6931_deepset}. With this intuition, we aim to use the theory of sets to make better use of the frame features within each video clip.

\begin{figure*}[ht]
\centering
\subfigure[Clip distance metric]{
\includegraphics[width=0.21\linewidth]{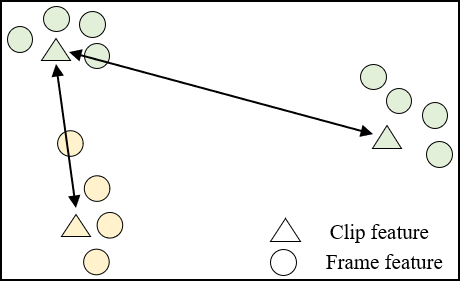}
\label{fig:c}
}
\hfill
\subfigure[Ordinary distance metric]{
\includegraphics[width=0.21\linewidth]{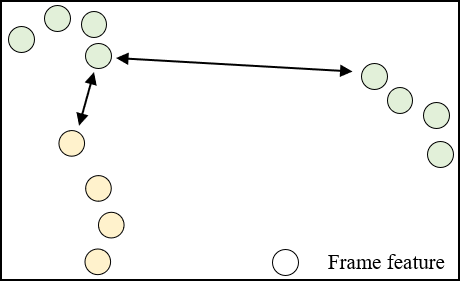}
\label{fig:o}
}
\hfill
\subfigure[Hausdorff distance metric]{
\includegraphics[width=0.21\linewidth]{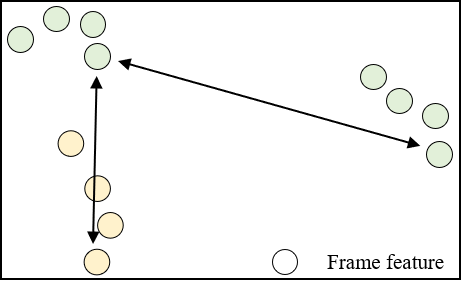}
\label{fig:H}
}
\hfill
\subfigure[Hybrid distance metric]{
\includegraphics[width=0.21\linewidth]{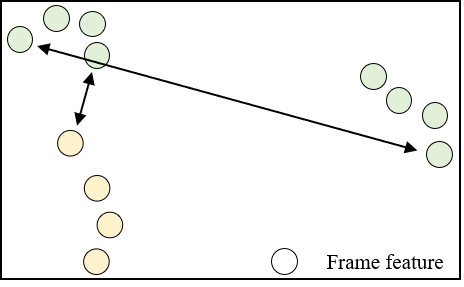}
\label{fig:HD}
}
\caption{Geometry interpretation of different distance metrics. (a), (b), (c), and (d) denote L$_2$ distance metric between clip representation, ordinary distance metric, Hausdorff distance metric, and hybrid distance metric between sets. The color represents the class of samples. }
\label{fig:metric}
\end{figure*}

In this paper, we propose to model the frame features within a clip as a set and propose to use the distance between sets in the triplet loss. Different from the L$_2$ distance between the aggregated clip features (see Fig.~\ref{fig:c}), the distance between sets considers every pair-wise distance in two sets and explores more information of the frame features. In set theory, the distance between sets is usually measured by ordinary distance (see Fig.~\ref{fig:o}) or Hausdorff distance (see Fig.~\ref{fig:H}). However, these set distance measures cannot fully utilize hard frames (\ie, hard positive and hard negative) in a triplet. To construct an effective set triplet loss, we further propose a hybrid distance metric (see Fig.~\ref{fig:HD}), where the hard frames for anchor-positive and anchor-negative sets are considered explicitly. In essence, our hybrid distance metric aims at penalizing the hard frames between sets (\ie, \textcolor{red}{$\leftrightarrow$} and \textcolor{blue}{$\leftrightarrow$} in Fig.~\ref{fig:example}). Fig.~\ref{fig:bar} shows the comparison of retrieval accuracies from video re-ID models, trained {\it without} our set-aware triplet loss, and {\it with} our set-aware triplet loss, across four video re-ID datasets. We further apply the class prototypes to frame-level features to construct hard sets by comparing the similarity between the class prototype and frame feature with the same instance. Then the constructed set acts as a hard positive set.


\noindent{\textbf{Contributions.}} The contributions of this work are summarized as follows:
\begin{itemize}
    \item We model the video clip as a set\footnote{In the remainder of this paper, we will use \say{clip} and \say{set} interchangeably}, and employ the distance metric between sets to construct the triplet loss. Furthermore, we propose a new hybrid set distance metric, which is tailored for the set triplet loss.

    \item We further model the weights in the last classification layer as class prototypes, to construct a hard positive set, \wrt each anchor set with the same identity.  
    
    \item Our algorithm achieves state-of-the-art performance across four standard video person re-ID datasets (\ie, {PRID-2011}~\cite{PRID}, {iLIDS-VID}~\cite{WangTaiqing2016TPAMIiiLIDS-VID}, {MARS}~\cite{ZhengLiang2014ECCVMAR} as well as {DukeMTMC-VideoReID}~\cite{WuYu2018CVPROneShotforVideoReID}), showing the effectiveness of the proposed set augmented triplet loss.
\end{itemize}

\section{Related Work}\label{sec:relatedwork}

\subsection{Sets}
The concept of modeling the training data as a set has appeared in many applications, \eg, point cloud classification~\cite{NIPS2017_6931_deepset}, image tagging~\cite{NIPS2017_6931_deepset}, object localization~\cite{ribera2019}~\etal. In general, the response of set functions is insensitive to the order of the elements in the set and the work in~\cite{NIPS2017_6931_deepset} studies the structure of such functions. The most popular function is the pooling operation (\ie, max pooling, average pooling) across the elements of its input. For example, deep Convolutional Neural Networks (CNNs) use pooling layers to summarize the features in a patch~\cite{He_2016_CVPR}. In the point cloud classification task~\cite{qi2016pointnet}, a non-linear function extracts the latent representation of point coordination and the pooling function further summarizes the features of objects. Attention using non-local connections also acts as a set function as the attention weights are produced by pairwise similarities of pixel features~\cite{WangXiaolong2017CVPRNonLocalNN}. In~\cite{ribera2019}, the locations of objects are estimated by training a detector which minimizes the set distance between the prediction and ground truth of objects.

\subsection{Metric Learning}
Deep metric learning aims to project images to a low dimensional embedding space, in which the images with similar semantics are clustered together~\cite{Suh_2019_CVPR, Soumava_Siamese_iccv_19, Pengfei_2019_ICCV}. The most popular paradigm is to employ the triplet loss to penalize the positive pair or negative pair or both of them within a triplet~\cite{SchroffFlorian2015CVPRFaceNet}. However, the possible number of triplets is exponential to the number of samples in a mini-batch, leading to a prohibitive computational cost. Much effort has gone into mining the triplets efficiently~\cite{HermansAlexander2017, Pengfei_2019_ICCV, Suh_2019_CVPR}. For example, the hard mining strategy only selects the hard positive and hard negative for an anchor sample~\cite{HermansAlexander2017}. However, a hard mining strategy often leads to getting caught in local minima during optimization ~\cite{HermansAlexander2017}; thus the semi-hard mining method is further proposed to make use of more negative pairs~\cite{Pengfei_2019_ICCV}. Beyond mining the triplets in a mini-batch, the work in~\cite{Suh_2019_CVPR} employs the class signatures to mine hard negative classes for an anchor class in the whole dataset.

\subsection{Person Re-identification}
Most popular solutions for person re-ID mainly focus on learning an appearance-discriminative representation~\cite{ZhengLiang2016arXivReIDPastPresentFuture}. In general, the person representation is often encoded by the holistic appearance feature~\cite{Mancs}, or part features~\cite{Sun_2018_ECCV} or both of them~\cite{Pengfei_2019_ICCV}. Beyond an image-based person re-ID, a video-based re-ID system can make use of additional temporal cues within a couple of frames, thereby encoding a robust video representation~\cite{Li_2019_ICCV,COSAM_2019_ICCV,RFA-Net}. Various temporal modeling methods have been studied extensively to effectively fuse the frame features to encode a discriminative and robust video representation. In~\cite{McLaughlin2016CVPRRNNforVideoReID,RFA-Net}, a clip-level person representation is modeled by average/max temporal pooling of frame-level features; thereafter, frame features are regressed by a Recurrent Neural Network (RNN), whose final hidden state encodes the representation of the target. The temporal modeling also utilizes the attention mechanism, which aggregates frame features according to individual importance~\cite{gao2018revisiting,Li_2019_ICCV}. In~\cite{gao2018revisiting}, individual importance is generated to aggregate frame features in a weighted sum fashion. 



In contrast to existing works, our work utilizes the original frame features by modeling the video clip as a set, and employ the distance between sets to optimize the hard frame features. In the remainder of this paper, we will present our set triplet loss and empirically verify the superior performance of the proposed method.

\section{Method}\label{sec:method}
\subsection{Set Theory Revisited}



A function $f(\cdot)$, which maps its domain $\mathcal{X}$ to its range $\mathcal{Y}$, is considered as a function of sets if it is permutation invariant to the order of elements in the input. In other words, given a set (\ie, $\Mat{X} = \{ \Vec{x}_1, \ldots, \Vec{x}_s\}$) as input, the function $f$ holds that $f(\Mat{X}) = f(\Mat{P}\Mat{X})$ for any permutation matrix $\Mat{P}$. In this case, the domain of $f(\cdot)$ is the power set of $\Mat{X}$, \ie, $\mathcal{X} = \wp(\Mat{X})$.

Let $(\Mat{X}, d)$ be a metric space. The distance between two nonempty sets $\Mat{A}$ and $\Mat{B}$ in $\wp(\Mat{X})$ (\ie $D: \wp(\Mat{X}) \setminus \emptyset \times \wp(\Mat{X}) \setminus \emptyset \to \mathbb{R}$) measures the similarity of two sets. The ordinary distance between sets (see Fig.~\ref{fig:o}) is defined as:   
\begin{equation}
\begin{split}
D^o(\Mat{A}, \Mat{B}) = \inf_{\Vec{a} \in \Mat{A}, \Vec{b} \in \Mat{B}} d(\Vec{a}, \Vec{b}),
\end{split}\label{eq:ordinary}
\end{equation}
where $\inf$ denotes the infimum function. The ordinary distance metric could be interpreted as the minimum pair-wise distance between two sets.   

Another well-known set distance metric is the Hausdorff distance, which is defined as:
\begin{equation}
\begin{split}
D^h(\Mat{A}, \Mat{B}) &= \max \big\{\sup_{\Vec{a}\in \Mat{A}}d(\Vec{a}, \Mat{B}), \sup_{\Vec{b}\in \Mat{B}}d(\Vec{b}, \Mat{A}) \big\}     \\
& = \max \big\{\sup_{\Vec{a}\in \Mat{A}} \inf_{\Vec{b} \in \Mat{B}} d(\Vec{a}, \Vec{b}),                      \sup_{\Vec{b}\in \Mat{B}} \inf_{\Vec{a} \in \Mat{A}} d(\Vec{a}, \Vec{b}) \big\},
\end{split}\label{eq:hausdorff}
\end{equation}
where $\sup$ represents the supremum function. As shown in Fig.~\ref{fig:H}, the geometrical interpretation of the Hausdorff distance can be understood as the greatest of all the distances from an element in one set to the closest element in the other set.


\subsection{Triplet Loss}
When training a deep video feature extractor, we first sample a mini-batch, which contains $P$ different classes and $K$ video clips for each class, with each video clip having $T$ frames.  The network first extracts the frame features, denoted by $\Mat{A}_i = \{\Vec{a}_{i1}, \ldots, \Vec{a}_{iT}\},i = 1, \ldots, PK$. Then the network aggregates the frame features to a clip feature as ${\Vec{\hat{a}}}_i = \mathrm{Agg}(\Mat{A}_i)$. Given an anchor clip representation ${\Mat{\hat{a}}}_i^a$, one possible triplet is formed as $\{{\Mat{\hat{a}}}_i^a, {\Mat{\hat{a}}}_i^p, {\Mat{\hat{a}}}_i^n \}$, where the positive pair (\ie, $\{{\Mat{\hat{a}}}_i^a, {\Mat{\hat{a}}}_i^p \}$) shares the same label, while the negative pair (\ie, $\{{\Mat{\hat{a}}}_i^a, {\Mat{\hat{a}}}_i^n \}$) does not. The triplet loss aims to penalize the triplet in which the distance between the positive pair is not sufficiently smaller than that between the negative pair. The triplet loss with hard triplet mining is given by
\begin{equation}\label{eq:triplet}
\mathcal{L}^{\mathrm{hm}}_{\mathrm{ctri}} = \frac{1}{PK} \sum^{PK}_{i=1} \max(0,  d_i({\Mat{\hat{a}}}_i^a, {\Mat{\hat{a}}}_i^p) - d_i({\Mat{\hat{a}}}_i^a, {\Mat{\hat{a}}}_i^n) + \eta), 
\end{equation}
where $\eta$ is a task-specific margin. Existing video re-ID machines~\cite{Fu2018STASA,gao2018revisiting} only optimize the clip representation (see Fig.~\ref{fig:c}) and it has never been considered to optimize the frame features within each video clip.



\subsection{Set-aware Triplet Loss}


The nature of the triplet loss is to penalize the positive pairs with a large distance and negative pairs with a small distance. It works well in image re-ID where the triplets are constructed from the image features. However, in video re-ID, the distance measure is hampered by the aggregation operation, as shown in Fig.~\ref{fig:example}. To overcome this issue, we directly enforce the constraint of the triplet loss on the frame features. We first model the frame features within a video clip as a set and employ set theory to calculate the distance between sets. Eqn.~\eqref{eq:ordinary} and Eqn.~\eqref{eq:hausdorff} formulate the commonly used set distance metrics. However, the geometry interpretation of Eqn.~\eqref{eq:ordinary} and Eqn.~\eqref{eq:hausdorff} (see Fig.~\ref{fig:o} and Fig.~\ref{fig:H}) indicates that those two distance metrics cannot distinguish the distances from the hard positive frames (\textcolor{blue}{$\leftrightarrow$} in Fig.~\ref{fig:example}) and hard negative frames (\textcolor{red}{$\leftrightarrow$} in Fig.~\ref{fig:example}) simultaneously. Thus, we further propose a hybrid distance metric tailored to the nature of the triplet loss. 


Given a triplet, \ie, $\{\Mat{A}^a, \Mat{A}^p, \Mat{A}^n \}$, the hybrid distance metric is defined using the anchor-positive distance and anchor-negative distance individually, as follows:   
\begin{equation}
\begin{split}
D^{hd+}(\Mat{A}^a, \Mat{A}^{p}) = \sup_{\Vec{a}^a \in \Mat{A}^a, \Vec{a}^{p} \in \Mat{A}^{p}} d(\Vec{a}^{a}, \Vec{a}^{p}),
\end{split}
\end{equation}
and
\begin{equation}
\begin{split}
D^{hd-}(\Mat{A}^{a}, \Mat{A}^{n}) = \inf_{\Vec{a}^a \in \Mat{A}^a, \Vec{a}^{n} \in \Mat{A}^{n}} d(\Vec{a}^{a}, \Vec{a}^{n}),
\end{split}
\end{equation}
where $D^{hd+}$ and $D^{hd-}$ denote the positive pair distance and negative pair distance, respectively. Fig.~\ref{fig:HD} shows the geometrical interpretation of the hybrid distance metric. This formulation allows the loss to penalize the hard frames in each set with the set-aware triplet loss:
\begin{equation}\label{eq:stri}
\mathcal{L}^{\mathrm{hm}}_{\mathrm{stri}} = \frac{1}{PK} \sum^{PK}_{i=1} \max(0, D^{hd+}_i - D^{hd-}_i + \eta). 
\end{equation}


\subsection{Hard Positive Set Construction}
The network is also supervised by a cross-entropy loss to minimize the within-class variance. Once the network aggregates the frame features to a clip feature as ${\Vec{\hat{a}}}_i = \mathrm{Agg}(\Mat{A}_i)$. A following fully connected (FC) layer, parameterized by $\Mat{W}$, is used to predict the identity of the video, normalized by the $\mathrm{softmax}$ function, as $\Vec{p} = \mathrm{softmax}(\Mat{W}^{\top} {\Vec{\hat{a}}}_i)$. A cross-entropy loss is employed to maximize the log likelihood of ${\Vec{\hat{a}}}_i$ with respect to its label $c$ as follows:  
\begin{equation}\label{eq:softmax-ce}
\mathcal{L}_{\text{ce}} = \frac{1}{PK}\sum_{i = 1}^{PK}-\mathrm{log}\big(p(y_i = c| {\Vec{\hat{a}}}_i)\big). 
\end{equation}
In Eqn.~\eqref{eq:softmax-ce}, it holds that $p(y_i = c|{\Vec{\hat{a}}}_i) \propto \Vec{w}_{c}^{\top}{\Vec{\hat{a}}}_i$. The optimization will maximize $p(y_i = c|{\Vec{\hat{a}}}_i)$, thereby maximizing the similarity between $\Vec{w}_c$ and ${\Vec{\hat{a}}}_i$. Thus $\Vec{w}_c$ can be understood as a prototype feature for the class $c$. Given $K$ sets containing the same class $c$ in one mini-batch, we can further approximate the probability of each frame feature belonging to its label as: $p(y_j = c| \Vec{a}_j), j = 1, \ldots, KT$. For each class, we continue to mine $T$ frame features  ${\Mat{\hat{A}}} = \{\Vec{a}_r: r \in \Vec{i}'\}$, where $\Vec{i}'$ satisfies 
\begin{equation}
\Vec{i}' =   \{ r : \argmin_{r = 1, \cdots, KT} p_r; ~\mathrm{s.t.} |\Vec{i}'| = T\},
\end{equation}
and this set is summarized to a set representation (\ie, ${\Vec{\hat{a}}} = \mathrm{Agg}({\Mat{\hat{A}}})$), acting as a hard positive with respect to the original set features $\{{\Vec{\hat{a}}}_1, \ldots {\Vec{\hat{a}}}_K\}$ in the batch, where ${\Vec{\hat{a}}}_i = \mathrm{Agg}(\Mat{A}_i)$. Finally, we could form hard positive pairs as $\{{\Vec{\hat{a}}}_i, {\Vec{\hat{a}}}\}, i = 1, \ldots, K$. The hard positive pairs are also minimized by the triplet loss. Besides the hard positive set, we mine a hard negative clip representation to form a valid triple loss, denoted by $\mathcal{L}_{\mathrm{ctri}}^{\mathrm{hpsc}}$. Algorithm \ref{alg:mining} summarizes the process of constructing hard positives.

\begin{algorithm}[!t]
  \caption{Hard Positive Set Construction}
  \label{alg:mining}
  \begin{algorithmic}[1]
    \Require 
        $K$: Number of sets;
        $T$: Number of frame features in each set with same class;
        $\Mat{A}_i = \{\Vec{a}_{i1}, \ldots, \Vec{a}_{iT}\}$: A set of frame features;
        ${\Vec{\hat{a}}}_i$: Set feature;
        $\Mat{W} = \{\Vec{w}_1, \ldots, \Vec{w}_n\}$: Class prototypes;
        $c$: Class of sets
    \Ensure
        $\{{\Vec{\hat{a}}}_i, {\Vec{\hat{a}}}\}, i = 1, \ldots, K$: Hard positive pairs
    \State Merging all sets with the same class: $\Mat{A} = \{\Mat{A}_1, \ldots, \Mat{A}_T\} = \{\Vec{a}_1, \ldots, \Vec{a}_{TK}\}$     
    
    \State Calculate the probability of predicting class $c$ for each frame: 
    \begin{align*}
    p(y_j = c| \Vec{a}_j) = \frac{\exp({\Vec{w}_c^{\top}\Vec{a}_j})}{\sum_{m=1}^n\exp(\Vec{w}_m^{\top}\Vec{a}_j)},~~~j = 1 \ldots TK    
    \end{align*}
        
    \State Pick $T$ frame features with the lowest probability, satisfying
    \begin{align*}
    \Vec{i}' =   \{r : \argmin_{r = 1, \cdots, KT} p_r; ~\mathrm{s.t.} |\Vec{i}'| = T\}
    \end{align*}
    
    \State Construct a hard positive set: ${\Mat{\hat{A}}} = \{\Vec{a}_r: r \in \Vec{i}'\}$
    
    \State Summarize to hard positive set feature: ${\Vec{\hat{a}}} = \mathrm{Agg}({\Mat{\hat{A}}})$
    
    \State Form hard positive pairs: $\{{\Vec{\hat{a}}}_i, {\Vec{\hat{a}}}\}, i = 1, \ldots, K$

  \end{algorithmic}
\end{algorithm}

\begin{figure*}[ht]
  \centering
  \includegraphics[width = 0.8\textwidth]{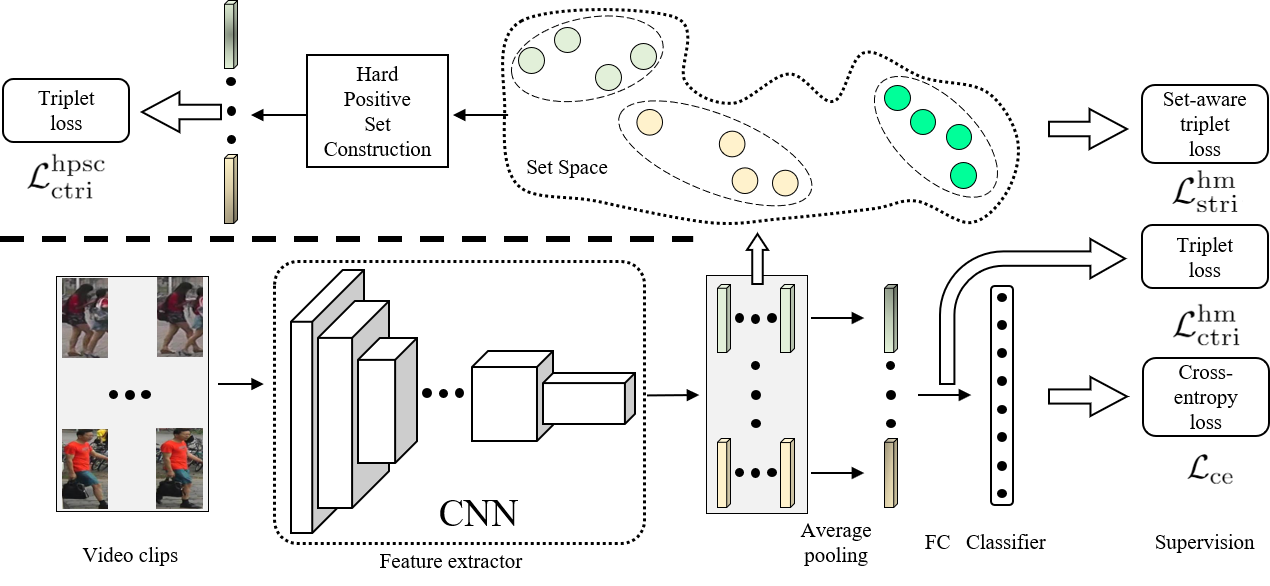}
  \caption{{The architecture of the proposed approach. The network receives frame images as input and produces the frame features.}}
  \label{fig:strucutre}
\end{figure*}

\subsection{Network and Optimization}
Fig.~\ref{fig:strucutre} shows the architecture of the deep network. The network receives a batch of video clips as input and produces frame representations. The original frame features are used to model the set and supervised by the set-aware triplet loss. We further use our proposed hard positive set construction to form hard positive pairs. Then average pooling is used to summarize the clip features. A vanilla triplet loss with hard mining and a triplet loss with hard positive set construction are utilized to supervise the clip features. An additional classifier is further used to train the network. The network is trained to update the parameters by jointly minimizing the multiple triplet losses and cross-entropy loss. The total loss function is formally formulated as:
\begin{equation}\label{eq:loss}
    \mathcal{L} = \lambda_1\mathcal{L}_{\mathrm{ce}} + \lambda_2\mathcal{L}_{\mathrm{ctri}}^{\mathrm{hm}} +
    \lambda_3\mathcal{L}_{\mathrm{ctri}}^{\mathrm{hpsc}} +
    \lambda_4\mathcal{L}_{\mathrm{stri}}^{\mathrm{hm}},
\end{equation}
where $\mathcal{L}_{\mathrm{ce}}$, $\mathcal{L}_{\mathrm{ctri}}^{\mathrm{hm}}$, $\mathcal{L}_{\mathrm{ctri}}^{\mathrm{hpsc}}$ and $\mathcal{L}_{\mathrm{stri}}^{\mathrm{hm}}$ denote cross entropy loss, clip-feature triplet loss with hard mining, clip-feature triplet loss with hard positive set construction, and set-aware triplet loss with hard mining. The loss terms are weighted by the factors $[\lambda_1, \lambda_2, \lambda_3, \lambda_4]$. 

\section{Experiments}\label{sec:expt}

\subsection{Datasets and Evaluation Protocol}

\begin{figure}[htb]
\centering
\includegraphics[width=0.9\linewidth]{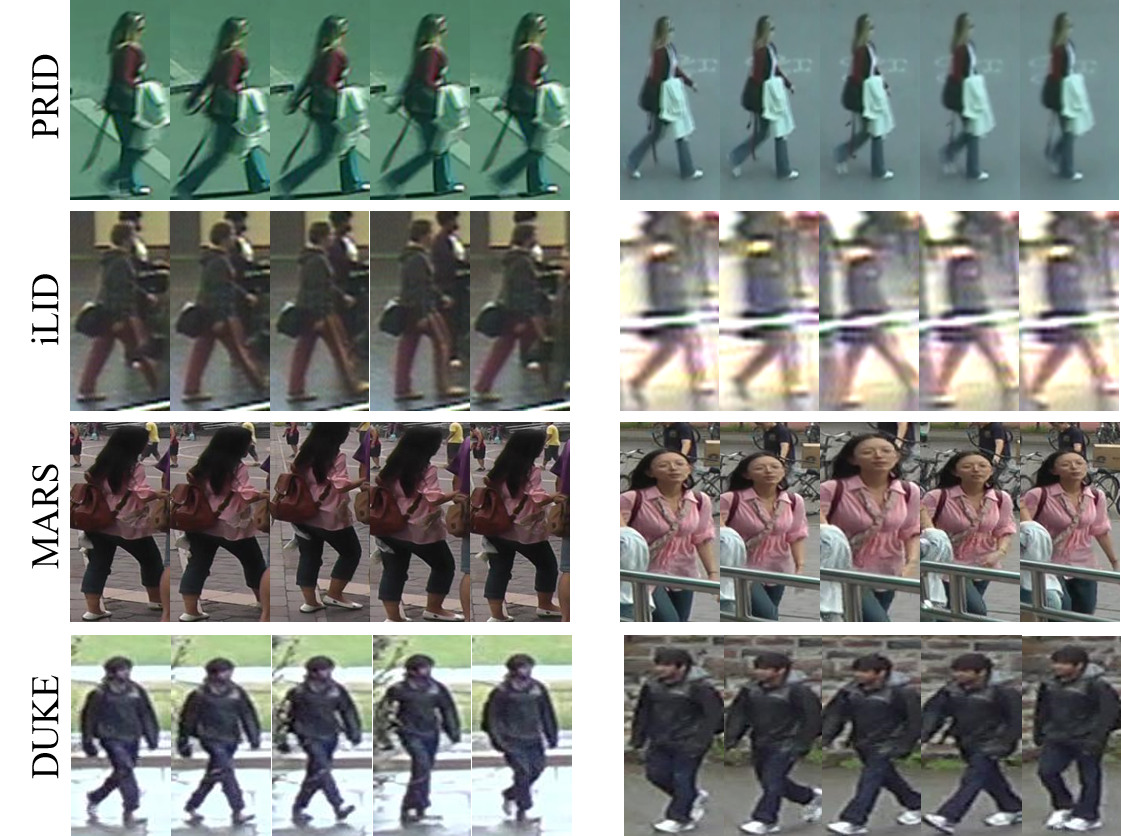}
\caption{Frames sampled from the pedestrian video sequences across four datasets. Each row shows two sequences of the same person captured by different cameras.}  \label{fig:sample}
\end{figure}

We evaluate our method on four popular video person re-identification benchmarks, including {PRID-2011}~\cite{PRID}, {iLIDS-VID}~\cite{WangTaiqing2016TPAMIiiLIDS-VID}, {MARS}~\cite{ZhengLiang2014ECCVMAR} and {DukeMTMC-VideoReID}~\cite{WuYu2018CVPROneShotforVideoReID}, with samples shown in Fig.~\ref{fig:sample}. The PRID-2011 consists of $200$ identities, each with $2$ video sequences, amounting to $400$ video sequences in total. Both the train and test sets contain $100$ person identities. The person trajectories are captured by two disjoint, static cameras. In each frame/image, the person bounding box is manually annotated. Similar to PRID-2011, iLIDS-VID is also a small scale dataset, which contains $600$ video sequences of $300$ identities, recorded by two cameras in an airport. Each of the train and test sets has $150$ person identities. The main challenge of this dataset is the occlusion of the target person. MARS is one of the large-scale video datasets. It has $1,261$ identities and $20,715$ video sequences captured by $6$ separate cameras. In this dataset, each video sequence is generated by the GMMCP tracker~\cite{GMMCP}, and the bounding box of each frame is automatically detected by DPM~\cite{DPM}. In this dataset, the train and test sets contain $631$ and $630$ person identities, respectively. The DukeMTMC-VideoReID is another large video re-ID dataset. This manually labeled dataset contains $702$ pedestrians for training, $702$ pedestrians for testing. Additionally, this dataset further employs $408$ extra pedestrians as distractors. Those $1812$ identities have $4832$ video sequences. Mean average precision (mAP) and cumulative matching characteristic (CMC) metrics are used to evaluate the proposed method. We report R-1, R-5, R-10 and R-20 values in the CMC metric. 

\begin{table*}[!ht]
\caption{Comparison with state-of-the-art approaches on PRID-2011, iLID-VID and MARS datasets. The $1^{\text{st}}$ best in \textbf{bold font}. $\dagger$ indicates the self-implemented network.}\label{table:sota-on-video-reid}
\begin{center}
\vspace{-0.3cm}
\scalebox{0.78}
{
\begin{tabular}{c|ccccc|ccccc|ccccc}
\hline
\multirow{2}{*}{{Methods}}
&\multicolumn{5}{c|}{\begin{tabular}[c]{@{}c@{}}PRID-2011\end{tabular}}
&\multicolumn{5}{c|}{\begin{tabular}[c]{@{}c@{}}iLIDS-VID\end{tabular}} 
&\multicolumn{5}{c}{\begin{tabular}[c]{@{}c@{}}MARS\end{tabular}}\\
\cline{3-5} \cline{8-10}  \cline{13-15}
&R-1&R-5&R-10 &R-20&mAP&R-1 &R-5&R-10&R-20&mAP&R-1&R-5&R-10&R-20&mAP \\
\hline
Chen~\etal~\cite{ChenDapeng2018CVPRVideoPersonReID}&88.6&99.1&-&-&90.9&79.8&91.8&-&-&82.6&81.2&92.1&-&-&69.4\\
+ Optcal flow&93.0&99.3&100.0&100.0&94.5&85.4&96.7&98.8&99.5&{87.8}&86.3&94.7&-&98.2&76.1
\\
QAN~\cite{liu_2017_qan}&90.3&98.2&99.3&100.0&-&68.0&86.8&-&97.4&-&73.7&84.9&-&91.6&51.7\\
Li~\etal~\cite{LiShuang2018CVPRSpatiotemporalAttentionforVideoReID}&93.2&-&-&-&-&80.2&-&-&-&-&82.3&-&-&-&65.8\\
PBR~\cite{Suh_2018_ECCV}&-&-&-&-&-&-&-&-&-&-&83.0&92.8&95.0&96.8&72.2\\
SCAN~\cite{ZhangRuimiao2018arXivSCANforVideoReID}&92.0&98.0&100.0&100.0&-&81.3&93.3&96.0&98.0&-&86.6&94.8&-&{98.1}&{76.7}\\
+ Optical flow &95.3&99.0&100.0&100.0&-&{88.0}&96.7&98.0&{100.0}&-&{87.2}&{95.2}&-&{98.1}&77.2\\
STIM-RRU~\cite{Liu2019SpatialAT}&92.7&98.8&-&99.8&-&84.3&96.8&-&{100.0}&-&84.4&93.2&-&96.3&72.7\\
COSAM~\cite{COSAM_2019_ICCV}&-&-&-&-&-&79.6 & 95.3 & - &-&- & 84.9 &95.5 &-&97.9 & 79.9\\
STAR+Optical flow ~\cite{BMVC2019STAR}&93.4&98.3&100.0&100.0&-&85.9&{97.1}&{98.9}&{99.7}&-&85.4&{95.4}&96.2&97.3&76.0\\
STA~\cite{Fu2018STASA}&-&-&-&-&-&-&-&-&-&-&86.3&95.7&-&98.1&80.8\\
VRSTC~\cite{Hou_2019_CVPR_VRSTC}&-&-&-&-&-&83.4&95.5&97.7&99.5&-&88.5&{96.5}&\textbf{97.4}&-&82.3\\
Zhao \etal~\cite{Zhao_2019_CVPR}&93.9&99.5&-&100.0&-&{86.3}&{97.4}&-&{99.7}&-&{87.0}&95.4&-&98.7&78.2\\
GLTR~\cite{Li_2019_ICCV}&95.5&100.0&-&-&-&86.0&98.0&-&-&-&87.0&95.7&-&98.2&78.4\\
MG-RAFA~\cite{Zhang_2020_CVPR}&95.9&99.7&-&100.0&-&\textbf{88.6}&98.0&-&99.7&-&88.8&97.0&-&98.5&\textbf{85.9}\\
STGCN~\cite{Yang_2020_CVPR}&-&-&-&-&-&-&-&-&-&-&\textbf{89.9}&96.4&-&98.3&83.7\\
\hline
ResNet-50&85.4&98.9&98.9&98.9&91.0&80.0&95.3&98.7&99.3&87.1&82.3&93.9&95.8&97.2&76.2\\
 + Set Triplet Loss (Ours) &90.2&99.6&100.0&100.0&93.6&85.3&96.0&98.6&99.4&90.4&85.3&95.4&97.1&98.2&81.8\\
 \hline
SE-ResNet-50&89.9&98.9&100.0&100.0&94.3&84.0&96.0&98.7&99.3&89.5&85.2&95.3&97.0&97.8&80.0\\
+ Set Triplet Loss (Ours) &\textbf{96.6}&\textbf{100.0}&\textbf{100.0}&\textbf{100.0}&\textbf{97.2}&\textbf{88.6}&\textbf{98.6}&98.7&\textbf{100.0}&\textbf{92.9}&87.9&\textbf{97.2}&97.1&\textbf{98.9}&83.2\\
\hline
GLTR$^{\dagger}$&94.4&99.7&100.0&100.0&95.3&85.2&96.7&97.3&99.7&91.1&86.4&95.4&96.9&97.7&78.8\\
+ Set Triplet Loss (Ours)&\textbf{96.6}&\textbf{100.0}&\textbf{100.0}&\textbf{100.0}&96.9&88.0&98.0&\textbf{99.3}&100.0&92.5&87.8&95.5&97.0&97.9&82.2\\
\hline
\end{tabular}
}
\end{center}
\end{table*}

\subsection{Implementation Details}
\subsubsection{Network and Data Organization}

We implement all experiments using the PyTorch~\cite{Pytorch} machine learning package. We use ResNet-50~\cite{He_2016_CVPR}, SE-ResNet-50~\cite{Hu_2018_CVPR} and GLTR \cite{Li_2019_ICCV} as baseline networks to evaluate our approach. Noted that the GLTR is self implemented version. All baselines are pre-trained on ImageNet~\cite{russakovsky2015imagenet}. The baseline network extracts each frame feature to the dimension of $2048$ and we further project them to a lower dimensional space of dimension $1024$. Thereafter, a set of frame features are fused to a clip-level video representation and a linear-transformation layer is further utilized to predict the class of the video representation. In each video clip, $T$ is chosen as $4$ in all experiments and $4$ frames are \textit{randomly} sampled from a video sequence. The frames are first resized to $288 \times 144$, and then randomly cropped to $256 \times 128$. The data augmentations used in our experiments include randomly flipping in the horizontal direction and random erasing (RE)~\cite{zhong2017random} during training. In the test phase, no data augmentation and re-ranking are used.


\subsubsection{Optimization Details}
We train the network using the Adam~\cite{kingma2014adam} optimizer with default momentum (\ie, $[\beta_1, \beta_2] = [0.9, 0.999]$). The learning rate is initialized to $3$e-$4$ for PRID-2011 and iLIDS-VID datasets, and $4$e-$4$ for MARS and DukeMTMC-VideoReID datasets. During training, the learning rate is decayed by a fixed factor of $1$e-$1$ at the $200^{\mathrm{th}}$ and $400^{\mathrm{th}}$ epoch for the PRID-2011 and iLIDS-VID , and the $100^{\mathrm{th}}$, $200^{\mathrm{th}}$ and $500^{\mathrm{th}}$ epoch for the MARS and DukeMTMC-VideoReID, respectively. The batch size is set to $16$ for the PRID-2011 and iLIDS-VID datasets and $32$ for the MARS and DukeMTMC-VideoReID datasets, respectively. In a mini-batch, both $P$ and $K$ are set to $4$ for the PRID-2011 and iLIDS-VID, whereas $P = 8$, $K = 4$ for the MARS and DukeMTMC-VideoReID. The margin in Eqn.~\eqref{eq:triplet} and Eqn.~\eqref{eq:stri}, \ie, $\eta$, is set to $0.3$ for all datasets. $[\lambda_1, \lambda_2, \lambda_3, \lambda_4] = [1, 0.5, 0.5, 0.5]$. In \textsection~\ref{sec:exp-ablation}, we will verify each loss component in the total loss function. We report the results of the network at its $800^{\mathrm{th}}$ epoch without any post processing tricks to boost the accuracy, \ie, re-ranking.

\subsection{Results}

We first compare our method to existing state-of-the art algorithms, as shown in Table~\ref{table:sota-on-video-reid} and Table~\ref{table:sota-on-video-reid_duke}.

\noindent{\textbf{Evaluation on PRID-2011.}} PRID-2011 is an old video re-ID dataset; thus only a few methods report the mAP value. To show the superiority of our method, we report both metrics for comparison in Table~\ref{table:sota-on-video-reid}. Our method outperforms MG-RAFA~\cite{Zhang_2020_CVPR} by $0.7\%$ on the R-1 value. Our approach also outperforms the state-of-the-art mAP value in~\cite{ChenDapeng2018CVPRVideoPersonReID} by $2.7\%$.

\noindent{\textbf{Evaluation on iLIDS-VID.}} Same as for the PRID-2011 dataset, we report the CMC accuracy and mAP value in Table~\ref{table:sota-on-video-reid}. On the iLIDS-VID dataset, our method also achieves state-of-the-art performance. In particular, our network has the same R-1 value with MG-RAFA~\cite{Zhang_2020_CVPR} and outperforms the state-of-the art mAP values by $5.1\%$ in~\cite{ChenDapeng2018CVPRVideoPersonReID}.

\noindent{\textbf{Evaluation on MARS.}} Compared with MG-RAFA~\cite{Zhang_2020_CVPR}, the state-of-the-art algorithm on the MARS dataset (see Table~\ref{table:sota-on-video-reid}), our method improves the R-5 and R-20 by $0.2\%$ and $0.4\%$  and achieves competitive performance on the R-1 and mAP value. 


\noindent{\textbf{Evaluation on DukeMTMC-VideoReID.}} We further evaluate our method on the DukeMTMC-VideoReID dataset. Table~\ref{table:sota-on-video-reid_duke} compares the performance between our network and existing state-of-the-art algorithms and demonstrates that our method outperforms the STGCN~\cite{Yang_2020_CVPR} by $0.2\%$ in mAP. Our methods also outperform STA~\cite{Fu2018STASA} by $0.2\%/0.8\%$ and GLTR by $0.1\%/2.0\%$ in R-1$/$mAP respectively.

\begin{table}[ht]
\caption{Comparison with the state-of-the-art approaches on the DukeMTMC dataset. The $1^{\text{st}}$ best in \textbf{bold font}. $\dagger$ indicates the self-implemented network.}\label{table:sota-on-video-reid_duke}
\begin{center}
\vspace{-0.3cm}
\scalebox{0.72}
{
\begin{tabular}{c|ccccc}
\hline
\multirow{2}{*}{{~~Methods}~~}&\multicolumn{5}{c}{\begin{tabular}[c]{@{}c@{}}~~~~~DukeMTMC-VideoReID~~~~~\end{tabular}}\\
\cline{3-5}
&~~R-1~~&~~R-5~~&~~R-10~~&~~R-20~~&~~mAP~~\\
\hline
ETAP-Net~\cite{WuYu2018CVPROneShotforVideoReID}&83.6&94.6&-&97.6&78.3\\
STAR+Optical flow~\cite{BMVC2019STAR}&94.0&99.0&99.3&99.7&93.4\\
VRSTC~\cite{Hou_2019_CVPR_VRSTC}&95.0&99.1&99.4&-&93.5\\
STA~\cite{Fu2018STASA}&96.2&99.3&-&99.7&94.9\\
GLTR~\cite{Li_2019_ICCV}&96.3&99.3&-&99.7&93.7\\
STGCN~\cite{Yang_2020_CVPR}&\textbf{97.3}&99.3&-&99.7&95.7\\
\hline
ResNet-50&87.5&96.5&97.2&98.3&86.2\\
+ Set Triplet Loss (Ours)&93.4&98.4&99.8&99.2&91.9\\
\hline
SE-ResNet-50&90.2&97.3&98.0&98.9&89.7\\
+ Set Triplet Loss (Ours)
&{96.8}&\textbf{99.4}&\textbf{99.9}&\textbf{99.9}&\textbf{95.9}\\
\hline
GLTR$^{\dagger}$&96.0&99.2&99.3&99.5&93.5\\
+ Set Triplet Loss (Ours)&97.1&\textbf{99.4}&99.8&\textbf{99.9}&95.4\\
\hline
\end{tabular}
}
\end{center}
\end{table}

\subsection{Ablation Study}
\label{sec:exp-ablation}
In this section, we will conduct extensive experiments to evaluate the effectiveness of each component in this work.

\noindent{\textbf{Effect of set-aware triplet loss.}}
We first evaluate the effectiveness of set-aware triplet loss with different set distance metrics. In this study, we use the SE-ResNet-50 as the backbone network and employ all three distance metrics for the set-aware triplet loss. As shown in Table~\ref{table:set-triplet}, the set-aware triplet loss indeed helps the network to learn a discriminative person description. Compared with the commonly-used set distance metrics (\ie, ordinary distance, Hausdorff distance), the proposed hybrid distance metric brings the largest performance gain, showing that the optimization to hard frames of anchor-positive pairs and anchor-negative leads the network to create a discriminative video representation.


\begin{table}[ht]
\caption{Effect of set-aware triplet loss across the iLIDS-VID and DukeMTMC-VideoReID datasets. SATL: set-aware triplet loss, $D^o$: ordinary distance, $D^h$: Hausdorff distance, $D^{hd}$: Hybrid distance.}
\begin{center}
\vspace{-0.3cm}
\scalebox{0.84}{
\begin{tabular}{c|cccc}
\hline
\multirow{2}{*}{Model} &\multicolumn{2}{c|}{~~iLIDS-VID~~} &\multicolumn{2}{c}{DukeMTMC-VideoReID}\\
\cline{2-5}
&~~R-1~~&~~mAP~~&~~R-1~~&~~mAP~~\\
\hline
SE-ResNet-50&84.0&89.5&90.2&89.7\\
\hline
SATL w/ $D^o$&86.8&90.6&92.8&91.7 \\ 
SATL w/ $D^h$&87.6&91.1&94.1&92.9 \\ 
SATL w/ $D^{hd}$&\textbf{88.3}&\textbf{91.9}&\textbf{94.9}&\textbf{93.7}\\ 
\hline
\end{tabular}
}
\end{center}\label{table:set-triplet}
\end{table}

\noindent{\textbf{Effect of hard positive set construction.}}
We continue to verify the effectiveness of our hard positive set construction method. We still use the SE-ResNet-50 as the backbone network. Table~\ref{table:hard-construction} shows that our network benefits from the hard positive set construction method across two datasets. A reasonable explanation for this improvement is that the hard positive sample helps the network minimize the intra-class variance, thereby improving the performance of the network.

\begin{table}[ht]
\caption{Effect of hard positive set construction across the iLIDS-VID and DukeMTMC-VideoReID datasets. HPSC: hard positive set construction.}
\begin{center}
\vspace{-0.3cm}
\scalebox{0.84}{
\begin{tabular}{c|cccc}
\hline
\multirow{2}{*}{Model} &\multicolumn{2}{c|}{~~iLIDS-VID~~} &\multicolumn{2}{c}{DukeMTMC-VideoReID}\\
\cline{2-5}
&~~R-1~~&~~mAP~~&~~R-1~~&~~mAP~~\\
\hline
SE-ResNet-50&84.0&89.5&90.2&89.7\\
\hline
HPSC&\textbf{86.2}&\textbf{91.4}&\textbf{92.4}&\textbf{91.9}\\
\hline
\end{tabular}
}
\end{center}\label{table:hard-construction}
\end{table}

\begin{figure*}[ht]
  \centering
  \includegraphics[width = 0.86\textwidth]{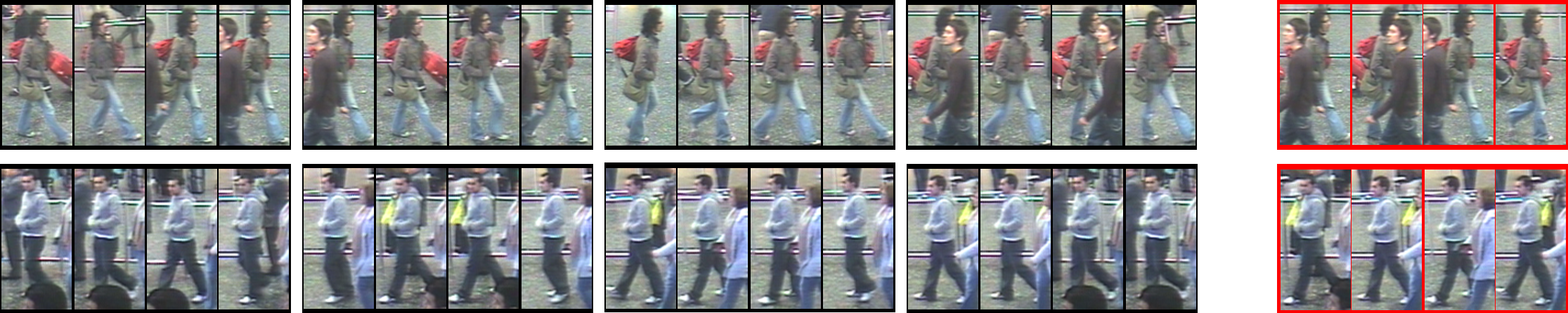}
  \caption{Example of hard positive set construction via Algorithm \ref{alg:mining} on the iLIDS-VID dataset. The original and constructed video clips/sets are framed by black and red lines, respectively. The constructed clip indicates that the frames with occlusions or distractors will be easily selected as hard samples by our algorithm. Images are sampled from two video sequences from different pedestrians. }\label{fig:hpsc}
\end{figure*}


\noindent{\textbf{Effect of each loss component.}}
In the study above, we have shown that our network achieves a performance gain from the set-aware triplet loss and the hard positive set construction method. In this study, we will verify each component in the total loss function. SE-ResNet-50 is also used here as the backbone network. The total loss function has four components (\ie, $\mathcal{L}_{\mathrm{ce}}$, $\mathcal{L}_{\mathrm{ctri}}^{\mathrm{hm}}$, $\mathcal{L}_{\mathrm{ctri}}^{\mathrm{hpsc}}$ and $\mathcal{L}_{\mathrm{stri}}^{\mathrm{hm}}$). Table~\ref{table:hard-construction} shows the effectiveness of each loss term. In this study, the baseline model is trained by cross-entropy loss (\ie, (\romannumeral1)). The rows in (\romannumeral2), (\romannumeral3), and (\romannumeral4) show that each of the triple losses provides complementary cues to optimize the network. In addition, the terms $\mathcal{L}_{\mathrm{ctri}}^{\mathrm{hpsc}}$ and $\mathcal{L}_{\mathrm{stri}}^{\mathrm{hm}}$ will further improve the performance of the network. In summary, this study reveals that our method helps the network to learn complementary information when encoding the person representation.  

\begin{table}[!ht]
\caption{Effect of each loss component across the iLIDS-VID and DukeMTMC-VideoReID datasets. $[\lambda_1, \lambda_2, \lambda_3, \lambda_4]$  denote the weights assigned to each loss term in Eqn.~\eqref{eq:loss}.}
\begin{center}
\vspace{-0.3cm}
\scalebox{0.8}{
\begin{tabular}{c|c|cc|cc}
\hline
\multicolumn{2}{c|}{\multirow{2}{*}{$[\lambda_1, \lambda_2, \lambda_3, \lambda_4]$}}  & \multicolumn{2}{c|}{~~iLIDS-VID~~}& \multicolumn{2}{c}{DukeMTMC-VideoReID}\\
\cline{3-6}
\multicolumn{2}{c|}{}&R-1&mAP&R-1&mAP\\
\hline
(\romannumeral1) &$[1, 0, 0, 0]$&74.7&82.5&80.2&79.6\\

(\romannumeral2) &$[1, 0.5, 0, 0]$&84.0&89.5&90.2&89.7\\

(\romannumeral3) &$[1, 0, 0.5, 0]$&82.0&87.6&87.3&85.2\\

(\romannumeral4) &$[1, 0, 0, 0.5]$&84.7&88.9&89.2&88.3\\

(\romannumeral5) &$[1, 0.5, 0.5, 0]$&85.2&90.4&91.4&90.9\\
(\romannumeral6) &$[1, 0.5, 0.5, 0.5]$&\textbf{89.3}&\textbf{92.9}&\textbf{96.8}&\textbf{95.9}\\ 
\hline
\end{tabular}
}
\end{center}\label{table:loss}
\end{table}






\noindent{\textbf{Visualization of hard positive set construction.}}
We further visualize the hard positive set construction by Algorithm \ref{alg:mining} on the iLIDS-VID dataset. The original and constructed video clips/sets are framed by black and red lines, respectively. As shown in Fig.~\ref{fig:hpsc}, we can observe that the frames with occlusions or distractors will be easily selected as hard samples by our algorithm. This observation is also in line with our intuition that the hard set is constructed from the hard frames in a batch.

\begin{figure}[ht]
\centering
	\subfigure[]{\includegraphics[width=0.4\linewidth, height = 4.8cm]{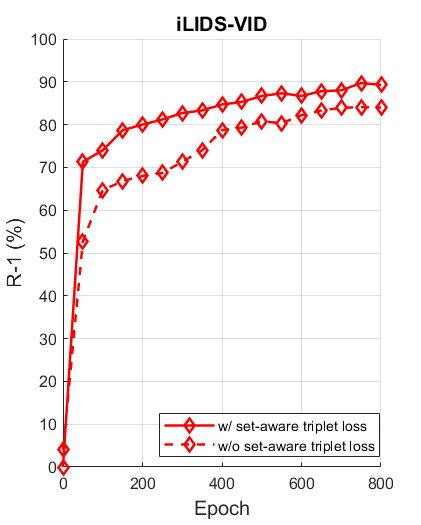}\label{fig:R1}}%
	\hfil
	\subfigure[]{\includegraphics[width=0.4\linewidth, height = 4.8cm]{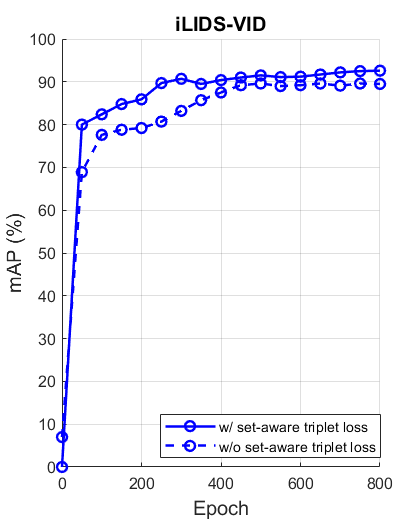}\label{fig:mAP}}%
	\hfil
	\caption{The training process of the network without set-aware triplet loss and with set-aware triplet loss on the iLIDS-VID dataset. (a): The R-1 value along the training process. (b): The mAP value along the training process.}\label{fig:curves}
\end{figure} 

\begin{figure}[ht]
\centering
	\subfigure[]{\includegraphics[width=0.45\linewidth]{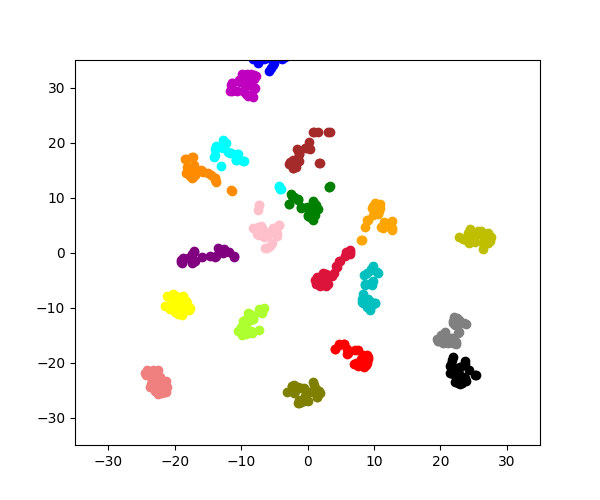}\label{fig:noset}}%
	\hfil
	\subfigure[]{\includegraphics[width=0.45\linewidth]{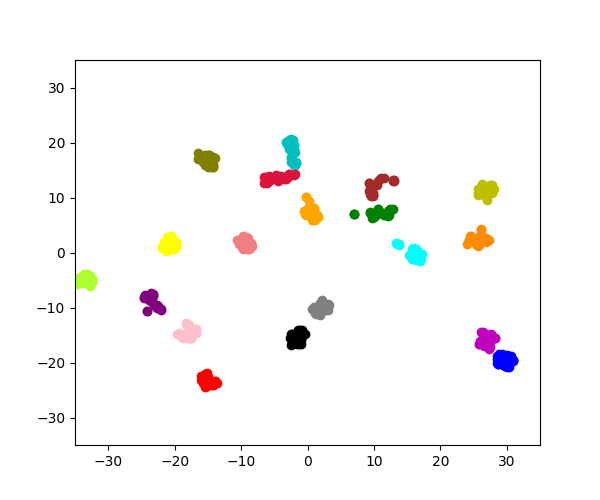}\label{fig:set}}%
	\hfil
	\caption{T-SNE visualization~\cite{vanDerMaaten2008} of learned features by the network (a) w/o set-aware triplet loss and (b) w/ set-aware triplet loss on the iLIDS-VID dataset. We select 20 people from the query set and visualize the frame features. Points with the same color denote the features of the same person. (Best viewed in color)}\label{fig:tsne}
\end{figure}

\noindent{\textbf{Training convergence and feature embedding.}}
In this part, we continue to demonstrate the superior performance of set-aware triplets by studying the training convergence and feature embedding of networks. In this study, we also use SE-ResNet-50 as the baseline network. Fig.~\ref{fig:R1} and Fig.~\ref{fig:mAP} show the training curves of the network with our set-aware triplet loss and without our set-aware triplet loss \wrt the R-1 value and mAP value respectively. Fig.~\ref{fig:noset} and Fig.~\ref{fig:set} visualize the features extracted by the network, trained without set-aware triplet loss, and with set-aware triplet loss. Both figures clearly show that the set-aware triplet loss indeed helps the network to learn a discriminative embedding space, in which the within-class variance is minimized and the between-class variance is maximized jointly.

\section{Conclusion}\label{sec:conclusion}
In this paper, we construct a triplet loss to optimize the frame features of the video person re-ID task, by modeling the video clip as a set. We employ the commonly-used distance metric to measure the distance between sets, \ie, ordinary distance and Hausdorff distance. Considering the hard pairs in the triplets, we further propose a new hybrid distance metric, which is defined for the anchor-positive pair and the anchor-negative pair separately. In addition, we also propose a hard positive set construction algorithm 
to decrease the within-class variance. Extensive experiments are conducted to verify the superior performance of the proposed method across the standard video person re-ID datasets.      

Future work includes employing the set distances to other general metric learning applications or other video-related applications.

\clearpage
{\small
\bibliographystyle{ieee_fullname}
\bibliography{egbib}
}

\end{document}